\documentclass[lettersize,journal]{IEEEtran}
\usepackage{amsmath,amsfonts}
\usepackage{algorithmic}
\usepackage{algorithm}
\usepackage{array}
\usepackage[caption=false,font=normalsize,labelfont=sf,textfont=sf]{subfig}
\usepackage{textcomp}
\usepackage{stfloats}
\usepackage{url}
\usepackage{verbatim}
\usepackage{graphicx}
\usepackage{adjustbox}
\usepackage{cite}
\usepackage{hyperref}
\usepackage{ulem}
\usepackage{xcolor}
\usepackage{booktabs}
\usepackage{multirow}
\usepackage{tabularx}
\usepackage{pifont}
\usepackage[switch]{lineno}

\hypersetup{
    colorlinks=true,
    citecolor=blue,
    linkcolor=blue,
    urlcolor=blue
}

\begin{document}

\title{Graph Foundation Models Can Become Effective Multivariate Time Series Anomaly Detectors}

\author{Zhuangyan Hu, Zhuoxuan Liang, Wei Li,~\IEEEmembership{Member,~IEEE}, Xiaohua Jia,~\IEEEmembership{Fellow,~IEEE}, Moustafa Youssef,~\IEEEmembership{Fellow,~IEEE}, Sumi Helal,~\IEEEmembership{Fellow,~IEEE}
\thanks{This work was supported by the National Science and Technology Major Project under Grants 2026ZD0127300 and 2026ZD0127303, and by the Supporting Fund of Intelligent Internet of Things and Crowd Computing under Grant B25029.}
\thanks{Zhuangyan Hu and Wei Li are with the College of Computer Science and Technology, Harbin Engineering University, Harbin 150001, China (e-mail: \{zhuangyan.hu,wei.li\}@hrbeu.edu.cn). 

Zhuoxuan Liang is with the Department of Computer Science, University of Exeter, Exeter, U.K. (e-mail: zl655@exeter.ac.uk). 

Xiaohua Jia is with the Department of Computer Science, City University of Hong Kong, Hong Kong, China (e-mail: csjia@cityu.edu.hk). 

Moustafa Youssef is with the Department of Computer Science and Engineering, The American University in Cairo, Egypt (e-mail: moustafa-youssef@aucegypt.edu). 

Sumi Helal is with the Department of Computer Science and Engineering, University of Bologna, Italy (e-mail: sumi.helal@unibo.it). 

\textit{Zhuangyan Hu and Zhuoxuan Liang contributed equally to this research. Corresponding authors: Wei Li.}}
}

\markboth{Journal of \LaTeX\ Class Files,~Vol.~14, No.~8, August~2021}%
{Shell \MakeLowercase{\textit{et al.}}: A Sample Article Using IEEEtran.cls for IEEE Journals}


\maketitle
\begin{abstract}
Multivariate time series anomaly detection (MTSAD) relies on modeling both temporal dynamics and inter-sensor dependencies to identify complex system faults. Although graph-based detectors explicitly model sensor topologies, they typically train task-specific networks from scratch on limited target-domain data, remaining vulnerable to spurious correlations and unable to reuse transferable relational priors. While pretrained Graph Foundation Models (GFMs) offer transferable topological priors and anomaly discriminability, directly adapting them to MTSAD is hindered by cross-modal feature-topology mismatch, unreliable pseudo-anomaly supervision, and decision distortion during node-to-window aggregation. To fill these gaps, we propose a novel adaptation framework, \textsc{Gfm4Mtsad}, which unleashes pretrained GFMs for unsupervised MTSAD with three key innovations. \ding{172} We introduce the Structure-and-Information Contract Alignment (SICA) module to bridge continuous time series and the pretrained GFM space. This module harmonizes principal subspace token projection, normal-prior-guided graph construction, and residual recovery to suppress spurious edges while retaining fine-grained temporal details. \ding{173} We propose a Mechanism-Aligned Pseudo-Anomaly Screening strategy (MAPS) to steer unsupervised learning. It applies localized dependency breaks while preserving empirical residuals, selecting hard negatives near the normal boundary for targeted supervision. \ding{174} We develop a Tail-Gated Evidence Calibration mechanism (TGEC) to eliminate node-to-window aggregation distortion. It calibrates multi-granular graph responses against normal operational references, strictly preserving benign score rankings while selectively refining extreme anomaly candidates in the tail distribution for reliable window detection and interpretable sensor evidence. Extensive evaluations across three real-world datasets with two GFM backbones demonstrate that \textsc{Gfm4Mtsad} achieves superior detection performance over existing baselines.
\end{abstract}


\section{Introduction}
\IEEEPARstart{M}{ultivariate} time series anomaly detection (MTSAD) safeguards mission-critical systems across industrial control \cite{mathur2016swat,rieth2017issues}, healthcare \cite{goldberger2000physiobank,yang2023deep}, and finance \cite{fahrmann2024anomaly,sun2026state,kim2026candi}, where missed faults risk catastrophic failures \cite{huang2025graph} or safety hazards \cite{sun2026state,kim2026candi}. Given scarce real-world anomalies and prohibitive annotation costs, MTSAD typically operates under an unsupervised regime trained solely on normal observations \cite{jang2025tail}. Moreover, anomalies manifest not only as isolated point deviations, but also as subtle breakdowns in temporal trajectories \cite{liu2025gcad} and cross-sensor correlations \cite{sun2026state}, demanding intricate spatio-temporal modeling to deliver both reliable window-level detection and fine-grained sensor localization \cite{liu2025large}.

Existing unsupervised detectors largely rely on temporal proxy objectives, such as reconstruction errors (e.g., USAD \cite{audibert2020usad}, OmniAnomaly \cite{su2019robust}) or pattern sensitivity learning \cite{li2025tsinr,feng2024sensitivehue}, yet fail to capture breakdowns in inter-sensor structural constraints. Graph-centric paradigms address this by modeling series as sensor graphs via Graph Neural Networks (GNNs, e.g., GDN \cite{deng2021graph}, CGSTA \cite{qi2026cgsta}, LaGraph \cite{zeng2026lagraph}) or causality-aware contrastive perturbations (e.g., CAROTS \cite{kim2025causality}). Nevertheless, these models ubiquitously follow a \textit{train-from-scratch} regime on limited target-domain data, leaving learned topologies susceptible to spurious edges and unable to transfer relational priors across systems. Coherently translating node-level responses into calibrated window decisions \cite{jang2025tail} for dependable fault attribution \cite{jeong2026pgrf} also remains underexplored.

Recently, pretrained Graph Foundation Models (GFMs, e.g., AnomalyGFM \cite{qiao2025anomalygfm}, OwlEye \cite{zheng2026owleye}) have offered transferable topological priors and intrinsic node-level anomaly discriminability learned from multi-domain graphs. However, a fundamental modality rift separates continuous temporal dynamics from static graph spaces: naive adaptation fails to activate pretrained priors \cite{wang2024gft}, amplifies structural noise \cite{zeng2026lagraph}, and discards temporal anomaly cues \cite{zheng2026owleye}. This raises a \textbf{key research question}: \textit{how to synergistically align the temporal dynamics and topological structures of multivariate time series (MTS) with pretrained GFMs?}

Yet, achieving such alignment in practical MTSAD extends beyond standalone feature projection; it requires an end-to-end adaptation pipeline that bridges dynamic temporal-topological inputs into the GFM space, steers boundary-aware discrimination without ground-truth labels, and translates fine-grained node responses into calibrated window decisions. Along this pipeline, we confront three intertwined challenges, as illustrated in Fig. \ref{fig:intro} (left):

\begin{figure*}[t]
    \centering
    \begin{minipage}[c]{0.62\textwidth}
        \centering
        \includegraphics[
            width=\linewidth,
            keepaspectratio
        ]{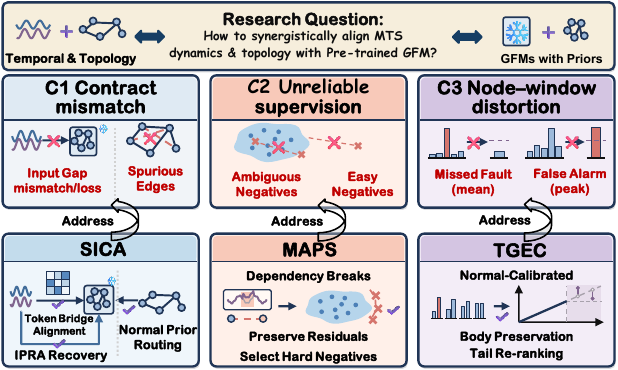}
    \end{minipage}
    \hfill
    \begin{minipage}[c]{0.36\textwidth}
        \centering
        \includegraphics[
            width=\linewidth,
            keepaspectratio
        ]{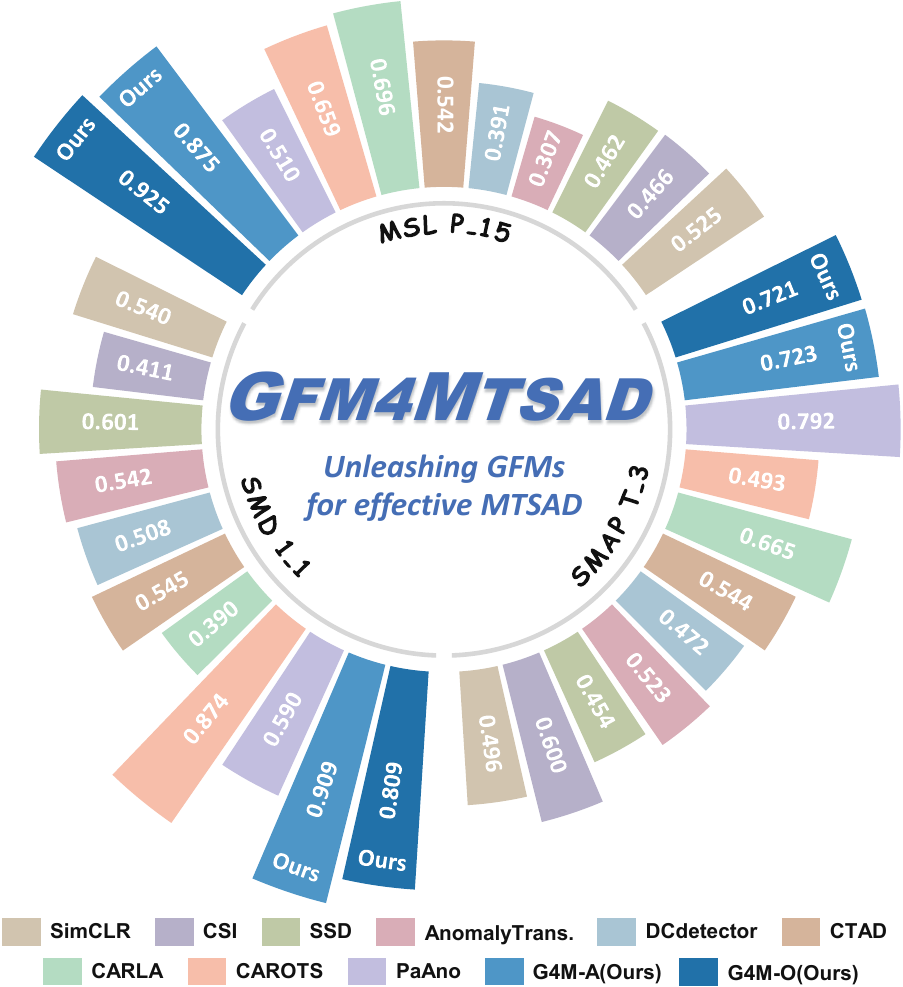}
    \end{minipage}

    \caption{\textbf{Left:} Three key challenges in adapting pretrained GFMs to MTSAD and their solutions: (C1) Contract mismatch, addressed by SICA through representation alignment and normal prior routing; (C2) Unreliable supervision, addressed by MAPS through dependency disruption and hard-negative selection; and (C3) Node-window distortion, addressed by TGEC through normality-guided tail calibration. \textbf{Right:} AUROC comparison across three benchmarks against nine representative baselines, where G4M-A and G4M-O instantiate \textsc{Gfm4Mtsad} with pretrained AnomalyGFM and OwlEye backbones.}
    \label{fig:intro}
\end{figure*}

\textit{C1: Contractual Mismatch Between Dynamic MTS and Pretrained GFMs.} Bridging continuous time series and discrete GFMs faces a dual mismatch in feature representations and graph topologies. On the feature side, high-dimensional dynamic embeddings from temporal frontends exhibit an inherent \textit{input gap} with the native token space of GFMs; crude linear projection or compression inevitably obliterates fine-grained temporal anomaly cues \cite{zheng2026owleye,wang2024gft}. On the topological side, empirically constructed dynamic graphs are plagued by noise \cite{huang2025graph} and \textit{spurious edges} \cite{feng2024sensitivehue}. Feeding such uncalibrated topologies into a pretrained GFM triggers noise amplification and over-smoothing during message passing \cite{zeng2026lagraph}, corrupting its relational priors. Establishing a structure-and-information alignment contract while preserving pretrained GFM knowledge thus constitutes the primary bottleneck for effective temporal-topological adaptation.

\textit{C2: Unreliable Supervision from Mechanism-Agnostic Pseudo-Anomalies.} Without ground-truth labels, discriminative adaptation heavily relies on synthesized pseudo-anomalies. Existing augmentation heuristics typically resort to random noise injection or global feature shuffling, which are detached from the spatio-temporal physical mechanisms of real-world systems \cite{kim2025causality,liu2025gcad}. Consequently, these perturbations either yield \textit{ambiguous negatives} that overlap with normal fluctuations and blur decision boundaries, or generate trivial \textit{easy negatives} lying far outside the normal support domain. Neither provides valid discriminative directions to activate the GFM. Effective adaptation therefore demands mechanism-aligned hard negatives that explicitly disrupt local temporal continuity and cross-sensor synergy near the normality boundary while preserving empirical residuals.

\textit{C3: Node-to-Window Distortion in Multi-Granular Evidence Aggregation.} Because real-world faults are often sparse and baseline scales vary markedly across sensors \cite{huang2025graph} and heterogeneous evidence channels \cite{jeong2026pgrf,jang2025tail}, naive readout severely distorts anomaly evidence. Specifically, \textit{mean pooling} dilutes sparse fault signals into \textit{missed faults}, whereas \textit{peak pooling} magnifies benign local spikes into \textit{false alarms}. Calibrating multi-scale node responses against normal operational references is therefore essential for reliable window-level detection and fine-grained sensor localization \cite{liu2025large}.

To address these challenges, we propose \textsc{Gfm4Mtsad}, a structure-and-information synergistic alignment framework that unleashes pretrained GFMs for unsupervised MTSAD, as illustrated in Fig. \ref{fig:intro} (left):

\textit{Structure-and-Information Contract Alignment (SICA):} To tackle C1, we propose the SICA module to seamlessly bridge dynamic MTS and pretrained GFMs. 
Rather than forcing the pretrained backbone to digest raw, noisy time series directly, the core idea of SICA is to wrap the GFM with peripheral adapters—translating raw series into clean, GFM-compatible graph inputs beforehand and restoring lost temporal details afterward. 
Specifically, it establishes three synergistic contracts: 
(1) \textit{Feature Contract:} A differentiable temporal frontend extracts dynamic sensor embeddings and maps them into the GFM's native token space via continuous \textit{Token Bridge alignment}. 
(2) \textit{Structure Contract:} Stable, high-frequency topologies mined from normal data serve as structural priors, coupled with dynamic \textit{normal prior routing} that gates message passing to prune spurious edges and curb noise propagation. 
(3) \textit{Information Contract:} An \textit{Information-Preserving Residual Adapter (IPRA)} is placed at the GFM output boundary for residual recovery, compensating for fine-grained temporal dynamics via a gated pathway. 
Throughout the pipeline, only the peripheral adapters and detection head are tuned, fully retaining pretrained relational priors.

\textit{Mechanism-Aligned Pseudo-Anomaly Screening (MAPS):} To tackle C2, we propose the MAPS strategy to furnish reliable anomaly supervision without ground-truth labels. 
Motivated by the intuition that real-world faults manifest as subtle breakdowns of physical regularities rather than random white noise, the core idea of MAPS is to deliberately disrupt local temporal habits or neighbor collaborations on a few target sensors, retaining only realistic, near-boundary hard negatives to teach the GFM where the true anomaly boundary lies. 
Concretely, we formalize four classes of spatio-temporal \textit{dependency breaks} directly from normal series: \textit{intra-node primitives} (\textit{lag-break} and \textit{local-function-shift}), which dismantle historical temporal dependencies and local evolution patterns; and \textit{inter-node primitives} (\textit{neighbor-break} and \textit{relation-flip}), which disrupt local collaborative dependencies and invert relational orientations. While applying localized interventions, MAPS preserves empirical prediction residuals on surrounding contexts and screens synthesized candidates via joint constraints—enforcing boundary proximity, mechanism disruption efficacy, graph locality, and discriminative directionality. This design selectively retains near-boundary hard negatives without contaminating the global graph, supplying directed task-level supervision for the GFM.

\textit{Tail-Guided Evidence Calibration (TGEC):} To tackle C3, we propose the TGEC mechanism to reliably translate fine-grained node responses into trustworthy window-level alerts. 
The core idea of TGEC is to calibrate multi-granular anomaly clues against each sensor's historical normal range before aggregation, ensuring that subtle single-sensor faults are not washed out by global averaging while routine background fluctuations do not trigger false alarms. 
Specifically, TGEC decomposes the GFM's responses into four hierarchical evidence channels: direct evidence (node-level acute shifts), local evidence (neighborhood-aggregated patterns), relational evidence (edge-level imbalances), and context evidence (whole-graph global representations). By calibrating these channels against normal operational baselines, TGEC aligns heterogeneous responses onto a comparable scale. This calibrated design achieves both \textit{body preservation}—stabilizing benign rankings to suppress false alarms induced by isolated local spikes—and \textit{tail re-ranking}, which selectively refines extreme candidates in the tail distribution so localized anomalies are not diluted during global pooling. Ultimately, TGEC yields dependable window-level anomaly decisions alongside interpretable sensor-level fault attributions.

The contributions of this paper are as follows:
\begin{itemize}
    \item To the best of our knowledge, we pioneer the adaptation of pretrained GFMs to unsupervised MTSAD, proposing a novel adaptation that bridges static graph foundation priors and continuous temporal observations.
    \item We propose SICA to harmonize feature, structure, and information contracts via Token Bridge alignment, normal prior routing, and IPRA residual recovery, suppressing topological noise while retaining fine-grained temporal details for reliable adaptation to pretrained GFMs.
    \item We introduce MAPS and TGEC to form a closed-loop supervision and decision pipeline, synthesizing mechanism-aligned hard negatives via localized dependency breaks and calibrating multi-granular evidence for accurate window detection and sensor localization.
    \item Extensive evaluations across three public benchmarks with two representative GFM backbones demonstrate that \textsc{Gfm4Mtsad} achieves superior performance over existing baselines (Fig. \ref{fig:intro}, right), substantiating the effectiveness of unleashing pretrained GFMs for MTSAD.
\end{itemize}

The remainder of this paper is organized as follows. Section \ref{sec:related_work} reviews related work, Section \ref{sec:method} details the proposed \textsc{Gfm4Mtsad} framework, and Section \ref{sec:experiments} presents the experimental results. Finally, Section \ref{sec:conclusion_and_future_work} concludes the paper and outlines future directions.

\section{Related Work}
\label{sec:related_work}

\subsection{Graph-based Time Series Anomaly Detection}
Graph-based methods represent monitored variables as nodes and jointly model temporal dynamics and cross-variable dependencies by learning inter-variable connections. 
GDN \cite{deng2021graph} learns a dependency graph via sensor embeddings and utilizes graph attention to predict variable responses, identifying anomalies based on prediction deviations. 
Subsequent research has increasingly focused on the temporal evolution and stability of inter-variable relationships: 
GCAD \cite{liu2025gcad} extracts sparse Granger causality graphs from the gradients of nonlinear predictors to detect anomalies through shifts in dependency patterns; 
CGSTA \cite{qi2026cgsta} constructs multi-scale dynamic graphs and stabilizes relation representations through normal reference alignment and cross-scale contrastive learning.

Structural information is also utilized for supervision construction and anomaly decision-making. 
CAROTS \cite{kim2025causality} generates positive and negative views through augmentations that preserve or perturb causal relationships, thereby providing structure-aware supervision for contrastive learning. 
PGRF-Net \cite{jeong2026pgrf} combines dynamic relationship learning with normal prototype modeling, utilizing learnable gating to fuse prediction deviations, structural changes, prototype deviations, and local spike evidence. 
These works have promoted the development of relationship modeling, structured augmentations, and multi-evidence detection. 
\textit{\textsc{Gfm4Mtsad} bridges these time series task requirements with pretrained graph anomaly representations, jointly designing temporal adaptation, mechanism-guided supervision, and normal reference scoring under the condition of a frozen backbone.}

\subsection{Graph Foundation Models}
GFMs aim to learn comprehensively reusable structural and attribute representations across graphs to support diverse graph domains and downstream tasks. 
GFT \cite{wang2024gft} regards computation trees formed by message passing as transferable units, connecting cross-domain and cross-task graph representation learning via a shared tree vocabulary. 
For graph anomaly detection, AnomalyGFM \cite{qiao2025anomalygfm} explicitly aligns node-to-neighborhood representation residuals with normal and abnormal prototypes, supporting zero-shot cross-graph inference and prompt adaptation based on a few normal nodes. 
OwlEye \cite{zheng2026owleye} combines cross-domain feature alignment, a multi-domain normal pattern dictionary, and truncated attention reconstruction to learn transferable patterns of normal behavior, enabling effective anomaly detection on unseen graphs.

These studies provide a foundation for reusing pretrained graph anomaly knowledge. 
Facing MTS, \textit{\textsc{Gfm4Mtsad} focuses its adaptation on the interface between temporal windows and pretrained GFMs: it guides node inputs and variable topology alignment using normal data, replenishes information not preserved by projection via residual pathways, and further organizes node anomaly responses into window-level detection scores and auxiliary localization evidence.}

\begin{figure*}[t]
    \centering
    \includegraphics[width=\textwidth]{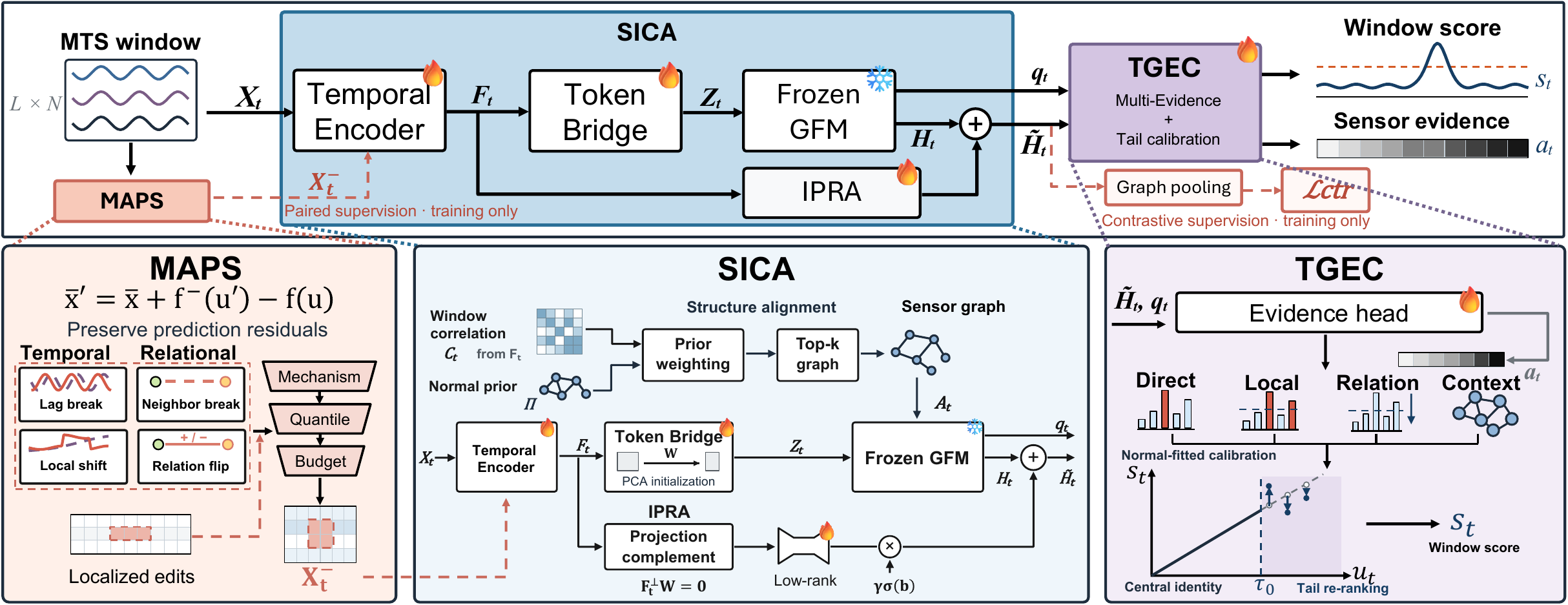}
    \caption{\textbf{Overall architecture of \textsc{Gfm4Mtsad}.} At a high level, \textsc{Gfm4Mtsad} adapts a frozen GFM to unsupervised MTSAD, outputting both window-level anomaly scores ($s_t$) and sensor-level localization evidence ($a_t$). Specifically, \textbf{(1) MAPS} (left, training only) synthesizes mechanism-aligned hard pseudo-anomalies ($X_t^-$) via residual-preserving interventions and multi-criteria screening to provide paired and contrastive supervision; \textbf{(2) SICA} (center) bridges the temporal-topological gap through prior-guided structure alignment, a Token Bridge, and orthogonal IPRA residual compensation while keeping the GFM backbone frozen; and \textbf{(3) TGEC} (right) integrates multi-granularity anomaly evidence via normal-calibrated tail re-ranking, strictly preserving central normal rankings. Flame and snowflake icons denote trainable modules and the frozen backbone, respectively.}
    \label{fig:framework}
\end{figure*}

\section{Method}
\label{sec:method}

\subsection{Problem Definition}
\label{method:subsec:preliminaries}

Let $\mathbf{X} = \{\mathbf{x}_t\}_{t=1}^T$ denote a MTS, where $\mathbf{x}_t \in \mathbb{R}^N$ represents the observations collected from $N$ sensors at time step $t$. A sliding window of length $L$ is denoted as $\mathbf{X}_t = [\mathbf{x}_{t-L+1}, \ldots, \mathbf{x}_t]^\top \in \mathbb{R}^{L \times N}$, whose $i$-th column, $\mathbf{X}_{t,:,i}$, corresponds to the observed trajectory of the $i$-th variable. The training set $\mathcal{D}_{\mathrm{N}}$ consists exclusively of normal windows. Given a pretrained GFM with frozen parameters $\boldsymbol{\theta}_0$, \textsc{Gfm4Mtsad} learns temporal adaptation and detection parameters to identify anomalous deviations in both temporal dynamics and cross-variable dependencies, outputting a window-level anomaly score $s_t$ along with auxiliary localization evidence $\mathbf{a}_t \in \mathbb{R}^N$ indexed by the sensors.

\subsection{\textsc{Gfm4Mtsad} Overview}
\label{method:subsec:overview}

The overall architecture of \textsc{Gfm4Mtsad} is illustrated in Fig.~\ref{fig:framework}. 
At a high level, \textsc{Gfm4Mtsad} repurposes pretrained GFMs (e.g., AnomalyGFM \cite{qiao2025anomalygfm}, OwlEye \cite{zheng2026owleye}) for MTSAD through three collaborative modules:

\paragraph{\textbf{SICA}} 
SICA serves as the core adaptation module that bridges temporal sensory streams with the pretrained GFM via three coordinated contracts: 
the \textit{Feature Contract} extracts decomposed temporal dynamics and anchors the Token Bridge projection within the dominant normal subspace; 
the \textit{Structure Contract} regularizes instantaneous correlations with global normal connectivity priors; and 
the \textit{Information Contract} recovers projection-excluded temporal features via an orthogonal residual bypass. 
This design aligns temporal-topological representations while preserving transferable relational priors across the pretrained backbone (Section~\ref{subsec:SICA}).

\paragraph{\textbf{MAPS}} 
MAPS is a mechanism-guided synthesis module that supplies informative hard normal-pseudo-anomaly pairs during unsupervised training. 
By executing residual-preserving local interventions on a sparse Graph Vector Autoregressive (Graph-VAR) reference and filtering candidates via dual quantile constraints, MAPS calibrates training difficulty near the normal boundary and translates edited coordinates into node-level paired supervision (Section~\ref{subsec:maps}).

\paragraph{\textbf{TGEC}} 
TGEC is a reference-guided post-processing calibrator that consolidates multi-granularity evidence for final scoring. 
By enforcing central identity across the normal body and restricting multi-evidence re-ranking strictly to upper-tail candidates, TGEC effectively sharpens tail anomaly detection while outputting standardized node responses for auxiliary sensor localization (Section~\ref{subsec:TGEC}).

\subsection{Structure-Information Contract Alignment (SICA)}
\label{subsec:SICA}

To address \textbf{C1}, SICA bridges continuous MTS and the strictly frozen GFM without eroding pretrained topological priors \cite{hu2021lora}. Directly feeding raw temporal windows into a frozen graph backbone faces three adaptation obstacles: standard linear projections struggle to disentangle composite trend-seasonal dynamics \cite{zeng2023dlinear} or exploit target-domain variance priors \cite{paischer2026parameter}; purely instantaneous window correlations contain spurious edges that trigger noise amplification and over-smoothing across frozen GNN layers \cite{chen2020measuring}; and dimensionality compression inevitably discards fine-grained temporal variations residing outside the projected subspace. Rather than forcing the frozen backbone to digest noisy temporal inputs, SICA wraps the GFM with peripheral adapters governed by three coordinated contracts (Fig.~\ref{fig:framework}): (i) a \textit{Feature Contract} that decomposes per-sensor temporal dynamics and anchors token projections within the dominant normal subspace; (ii) a \textit{Structure Contract} that regularizes transient window correlations with persistent normal connectivity priors to ensure reliable message passing; and (iii) an \textit{Information Contract} that recovers projection-excluded temporal details at the GFM output boundary via an orthogonal residual bypass.

\paragraph{Feature Contract}
To establish a data-informed token interface, we first isolate per-sensor trend-seasonal dynamics via a dedicated decomposition frontend \cite{zeng2023dlinear} and map them into the GFM token space: $\mathbf{F}_t=f_\phi(\mathbf{X}_t)$ and $\mathbf{Z}_t=\operatorname{Norm}_N(\mathbf{F}_t\mathbf{W})$. Here, $f_\phi$ is the learnable temporal frontend, and $\operatorname{Norm}_N$ normalizes each extracted feature across sensors. The trainable projection $\mathbf{W}$ is empirically initialized from the leading principal components of normal features, matching the GFM's native input dimension.

\paragraph{Structure Contract} 
Building on the temporal features $\mathbf{F}_t$, we construct a noise-resilient relational topology for message passing by regularizing instantaneous Pearson correlations $\mathbf{C}_t$ with global normal connectivity priors $\mathbf{\Pi}$:
\begin{equation}
\mathbf{A}_t
=
\mathcal{T}_{k_g}
\left(
\mathbf{C}_t\odot
[(1-\lambda_A)\mathbf{1}+\lambda_A\boldsymbol{\Pi}]
\right).
\label{eq:structure_contract}
\end{equation}
Here, $\mathbf{C}_t=|\operatorname{Corr}(\mathbf{F}_t)|$, and $\Pi_{ij}$ is the frequency with which edge $(i,j)$
appears in normal training graphs. The weight $\lambda_A$ controls the prior contribution; $\mathcal{T}_{k_g}$ retains the strongest $k_g$ neighbors per node, symmetrizes the graph by maximum, and adds self-loops. Graph construction is detached from gradient updates.The GFM produces $(\mathbf{H}_t,\mathbf{q}_t) =\mathcal{B}(\mathbf{Z}_t,\mathbf{A}_t)$, comprising node representations and native anomaly responses. Its backbone weights $\theta_0$ remain frozen, while native adaptation parameters $\psi$ may be trained. For hierarchical backbones, an optional reliability gate uses normal edge stability and the persistence of selected edges to mix the window graph with the identity matrix in the first layer and with the normal graph in subsequent layers.

\paragraph{Information Contract} 
The Token Bridge may discard anomaly-relevant variations outside its projection subspace. IPRA recovers these components through a gated low-rank residual:
\begin{equation}
\begin{aligned}
\mathbf{F}_t^\perp
&=
\mathbf{F}_t-
\mathbf{F}_t\operatorname{sg}(\mathbf{W}\mathbf{W}^+),\\
\widetilde{\mathbf{H}}_t
&=
\mathbf{H}_t+
\gamma\sigma(b)
\operatorname{GELU}
\left(\operatorname{RMS}(\mathbf{F}_t^\perp)\mathbf{U}\right)
\mathbf{V}.
\end{aligned}
\label{eq:ipra_update}
\end{equation}
Here, $\mathbf{W}^{+}$ is the pseudoinverse, $\operatorname{sg}$ stops gradients, and RMS normalization acts per sensor. The rank-$r$ bottleneck $\mathbf{U},\mathbf{V}$ is controlled by gate $\sigma(b)$ and fixed scale $\gamma$. We initialize $\mathbf{V}$ to zero and mean-pool $\widetilde{\mathbf{H}}_t$ to obtain $\widetilde{\mathbf{g}}_t$.

\subsection{Mechanism-Aligned Pseudo-Anomaly Screening (MAPS)}
\label{subsec:maps}

To address \textbf{C2}, MAPS synthesizes mechanism-aligned hard negatives with exact sensor-time coordinates to guide unsupervised learning. Unlike generative paradigms prone to mode collapse \cite{naiman2024generative}, costly iterative sampling \cite{lu2022dpm}, or auxiliary mask-alignment overhead \cite{jin2025dual}, and unlike heuristic augmentations (e.g., Gaussian jittering or masking) \cite{kim2025causality} that introduce high-frequency artifacts and trivial numerical shortcuts, MAPS conceptualizes multivariate anomalies as localized breakdowns in spatio-temporal couplings. Specifically, MAPS operates through a two-stage intervention-and-screening pipeline directly over normal series: first, it fits a sparse Graph-VAR reference to characterize normal cross-sensor dependencies and executes \textit{residual-preserving local interventions} across four structural primitives, disrupting temporal or relational regularities while retaining empirical background fluctuations; second, to prevent synthesized candidates from degenerating into either trivial far-off outliers or imperceptible normal shifts, it enforces \textit{quantile-constrained hard screening} to select minimal-edit candidates situated near the upper normal energy boundary. This provides informative window-level negatives alongside exact perturbed sensor sets for fine-grained localization supervision.

\paragraph{Residual-Preserving Local Interventions} 
To faithfully mimic structural faults on normal dynamics, we first characterize cross-sensor dependencies via a sparse Graph-VAR model \cite{liu2025gcad,tank2022neuralgranger}: $f_i(\mathbf{u}_{\tau,i}) =c_i+\mathbf{w}_i^\top\mathbf{u}_{\tau,i}$, where $\mathbf{u}_{\tau,i}$ collects lagged observations of sensor $i$ and its selected neighbors. The coefficients are fitted on normal data, and $\bar{x}_{\tau,i}$ denotes a normally standardized observation.

Within an intervention interval $\mathcal{I}_t$, we replace the predicted component while retaining the original residual:
\begin{equation}
\bar{x}'_{\tau,i}
=
\bar{x}_{\tau,i}
+
f_i^-(\mathbf{u}'_{\tau,i})
-
f_i(\mathbf{u}_{\tau,i}),
\qquad \tau\in\mathcal{I}_t.
\label{eq:maps_intervention}
\end{equation}
Here, $\mathbf{u}'_{\tau,i}$ contains the synthesized history. Let $d_{\tau,ij}$ denote neighbor $j$'s lagged contribution to $f_i$. We select $j^\star$ by the largest sum of lag-wise mean absolute contributions over $\mathcal{I}_t$. For strength $\alpha\in[0,1]$, the intervention is
\begin{equation}
f_i^-(\mathbf{u}_{\tau,i})
=
\begin{cases}
(1-\alpha)f_i(\mathbf{u}_{\tau,i})
+\alpha f_i(\mathbf{u}_{\tau-d,i}),
& \text{lag},\\
f_i(\mathbf{u}_{\tau,i})-\alpha d_{\tau,ij^\star},
& \text{neighbor},\\
f_i(\mathbf{u}_{\tau,i})-2\alpha d_{\tau,ij^\star},
& \text{relation},\\
f_i(\mathbf{u}_{\tau,i})+\alpha\nu_i,
& \text{local}.
\end{cases}
\label{eq:maps_primitives}
\end{equation}
The offset $d$ is a positive integer. We set $\nu_i$ to $-1$ when the mean normal prediction over $\mathcal{I}_t$ is nonnegative, and to $1$ otherwise. Relation flip reverses the selected coupling when $\alpha$ exceeds $0.5$. The intervened predictor and neighbor contributions are evaluated recursively on the synthesized history. 

\paragraph{Quantile-Constrained Hard Screening} 
We seek detectable mechanism changes under bounded edits. Let $\mathbf{r}_i(\mathbf{X})$ contain the Graph-VAR residuals standardized by normal residual statistics. For $n$ valid scoring coordinates, define $E_i(\mathbf{X})=\|\mathbf{r}_i(\mathbf{X})\|_2/\sqrt{n}$. Relative to the fixed normal window $\mathbf{X}_t$,
the paired energy of a candidate is
\begin{equation}
[E_i^-(\mathbf{X}')]^2
=
E_i(\mathbf{X}_t)^2
+
\frac{
\|\mathbf{r}_i(\mathbf{X}')-\mathbf{r}_i(\mathbf{X}_t)\|_2^2
}{n}.
\label{eq:maps_energy}
\end{equation}

Let $\mathcal{U}_t$ contain candidates satisfying finite-value, temporal-context, normal-support, and editing-locality checks. The normalized edit magnitude $\rho_i(\mathbf{X}')$ is the root-mean-square change over $\mathcal{I}_t$, scaled by the sensor's normal scale or valid-state span. Let $q_i$ and $q_i^{\mathrm u}$ denote the target and upper normal-energy quantiles, and $\sigma_i^E$ the normal-energy scale, estimated using the same scoring length as the candidate. We select
\begin{equation}
\begin{aligned}
\mathbf{X}_t^-
&=
\arg\min_{\mathbf{X}'\in\mathcal{U}_t}
\left[
\frac{|E_i^-(\mathbf{X}')-q_i|}{\sigma_i^E}
+\lambda_\rho\rho_i(\mathbf{X}')
\right],\\
\text{s.t.}\quad
&\rho_{\min}\le\rho_i(\mathbf{X}')\le\rho_{\max},\\
&E_i(\mathbf{X}_t)+\eta\sigma_i^E
\le E_i^-(\mathbf{X}')
\le q_i^{\mathrm u}.
\end{aligned}
\label{eq:maps_screen}
\end{equation}
The constraints require a mechanism change while bounding its energy and edit magnitude. The selected view and its directly edited sensor set $\mathcal{D}_t$ supervise training. If no candidate is feasible, the window contributes only to positive-view consistency.

\subsection{Tail-Gated Evidence Calibration (TGEC)}
\label{subsec:TGEC}

To address \textbf{C3}, TGEC eliminates decision distortion when translating fine-grained node responses into window-level anomaly scores and sensor attributions. In real-world faults, anomaly clues manifest across heterogeneous granularities—spanning isolated node spikes, neighborhood shifts, relational imbalances, and global context—whose raw scales and distributions differ substantially. However, naively pooling these channels or applying unconstrained global fusion \cite{jeong2026pgrf} easily dilutes sparse sensor faults or overfits benign background fluctuations across normal windows. Drawing inspiration from Extreme Value Theory (EVT) \cite{clifton2014extreme,siffer2017anomaly}, TGEC resolves this dilemma through a two-step calibration and selective fusion paradigm: first, it extracts four hierarchical anomaly descriptors and calibrates each onto a unified normal-reference tail scale to form a comparable multi-evidence consensus; second, it enforces \textit{central identity with tail gating}, which strictly preserves baseline rankings across the normal score body to prevent false alarms while activating multi-evidence re-ranking exclusively on upper-tail candidates, accompanied by standardized direct responses for sensor-level localization.

\paragraph{Multi-Granularity Evidence Aggregation}
Direct and local node heads add bounded, tanh-scaled MLP corrections to stabilized log-native responses, conditioned on the adapted node features and log-responses. Let $e_{t,i}$ denote the resulting direct response.

We extract four complementary window descriptors. \textit{Direct} evidence averages the largest $k_n$ direct responses; \textit{local} evidence averages local-head responses across sensors; \textit{relation} evidence measures the Euclidean norm of mean-centered local responses; and \textit{context} evidence combines the pooled graph representation with response statistics through an MLP.

Each descriptor is calibrated against its normal reference. Sigmoid-smoothed comparisons estimate its lower and upper tail masses, using a unit pseudocount and a denominator equal to the reference size plus one. We double the smaller tail mass, cap it at one, and take its negative logarithm. The consensus $v_t$ averages the largest $k_e$ calibrated values. The consensus is $v_t=\operatorname{TopMean}_{k_e} (\{\widetilde e_t^a\}_{a\in \{\mathrm d,\mathrm l,\mathrm r,\mathrm c\}})$.

\paragraph{Tail-Guided Re-Ranking} 
Per-sensor standardization of direct responses using normal means and stabilized standard deviations yields
the auxiliary evidence $\mathbf{a}_t$. Temperature-scaled log-mean-exp pooling of these responses produces the baseline window score $S_t$, with temperature $T_p$. 
We standardize $S_t$ using the median and scaled median absolute deviation of preceding scores only when fewer than half the nodes have valid normal Graph-VAR fits. Within each sequence, $b_t$ is the larger of the resulting score and its trailing mean over up to ten windows, including the current one. 
Given normal references $\{b_r^{\mathrm N}\}_{r=1}^{m_b}$, we set $\tau_0=-\log\alpha_0$ for tail mass $\alpha_0$ and compute
\begin{equation}
\begin{aligned}
u_t
&=
-\log
\frac{
1+\sum_{r=1}^{m_b}\mathbb{I}[b_r^{\mathrm N}\ge b_t]
}{m_b+1},\\
s_t
&=
\begin{cases}
u_t, & u_t<\tau_0,\\
\max\{\tau_0,\omega u_t+(1-\omega)v_t\},
& u_t\ge\tau_0.
\end{cases}
\end{aligned}
\label{eq:TGEC_rerank}
\end{equation}
The weight $\omega$ balances baseline and multi-evidence responses; only upper-tail candidates are re-ranked, while the remaining $u_t$ coordinates are unchanged.

\subsection{Training Objectives}
\label{subsec:training-objective}

\paragraph{Trainable Parameters}
Each normal window is paired with a positive view generated by subtle affine transformations and a MAPS pseudo-anomalous view. The trainable parameters $\Theta$ comprise the temporal frontend, Token Bridge, IPRA, direct evidence head, and native GFM adaptation parameters $\psi$. The GFM backbone, its running statistics, and the local and contextual heads remain frozen.

Below, $e_{t,i}$ denotes direct node evidence. We use $\mathbb{E}$ for averaging over normal windows and $\mathbb{E}_w$ for averaging with detached candidate weights. Each negative-pair loss uses windows with valid candidates and the nonempty sensor sets required by that loss; empty subsets contribute zero.

\paragraph{Spatially Constrained Node Localization}
Let $\mathcal{D}_t$ contain directly edited sensors. Their paired response increments are $\Delta_{t,i}=e_{t,i}^-
-\operatorname{sg}(e_{t,i})$. We rank these increments above those of other sensors:
\begin{equation}
\mathcal{L}_{\mathrm{node}}
=
\mathbb{E}_w
\left[
\operatorname{mean}_{i\in\mathcal{D}_t,\,
j\notin\mathcal{D}_t}
[m_n+\Delta_{t,j}-\Delta_{t,i}]_+
\right],
\label{eq:node_loss}
\end{equation}
where $m_n>0$ is the margin and $[z]_+=\max(z,0)$. Sensors outside $\mathcal{D}_t$ may include propagated effects.

\paragraph{Unified Optimization Objective}
The total objective combines window discrimination, sensor localization, and response consistency:
\begin{equation}
\begin{aligned}
\mathcal{L}
={}&
\mathcal{L}_{\mathrm{ctr}}
+\lambda_s
(\mathcal{L}_{\mathrm{rank}}
+\lambda_c\mathcal{L}_{\mathrm{cls}})\\
&+\lambda_n\mathcal{L}_{\mathrm{node}}
+\lambda_r(\mathcal{L}_++\mathcal{L}_{\mathrm{keep}}).
\end{aligned}
\label{eq:total_loss}
\end{equation}
The nonnegative weights control the contributions of the corresponding objectives.

The contrastive loss $\mathcal{L}_{\mathrm{ctr}}$ uses a two-logit softmax over cosine similarities scaled by temperature $T_c$, favoring the normal--positive pair over its paired normal--pseudo-anomalous pair. The window hinge loss $\mathcal{L}_{\mathrm{rank}}$ penalizes violations of $S_t^-\ge S_t+m_w$. The classification loss $\mathcal{L}_{\mathrm{cls}}$ averages binary cross-entropy over the three baseline-score logits with labels $(0,0,1)$. All training scores precede temporal buffering and TGEC re-ranking.

Both consistency terms use Smooth L1 with unit transition threshold and detached normal responses as targets.
The positive term $\mathcal{L}_+$ averages over all sensors and normal windows. The negative term $\mathcal{L}_{\mathrm{keep}}$ first averages over the preserved sensors $\mathcal{K}_t$ within each window, then uses the candidate-weighted window average. The set $\mathcal{K}_t$ excludes direct edits, propagated effects, and mechanism-based exclusions.

\section{Experiments}
\label{sec:experiments}

We conduct extensive experiments on three real-world MTSAD benchmarks across two heterogeneous GFM backbones to evaluate the effectiveness of \textsc{Gfm4Mtsad}. Specifically, we investigate: 
\begin{itemize}
    \item \textbf{RQ1:} How does \textsc{Gfm4Mtsad} achieve superior performance across multiple datasets compared with other MTSAD methods?
    \item \textbf{RQ2:} How effective are the individual components of \textsc{Gfm4Mtsad}?
    \item \textbf{RQ3:} How does the proposed tail-gated calibration mechanism operate to suppress false alarms while preserving normal rankings?
    \item \textbf{RQ4:} How do key parameters affect \textsc{Gfm4Mtsad}’s performance?
\end{itemize}

\subsection{Experimental Settings}
\label{subsec:experimental-settings}

\begin{table}[t]
    \centering
    \caption{Statistics of the Evaluation Datasets.}
    \label{tab:datasets}

    \footnotesize
    \renewcommand{\arraystretch}{1.12}
    \setlength{\tabcolsep}{3pt}

    \begin{tabular*}{\columnwidth}{
        @{\extracolsep{\fill}} lccccc @{}
    }
        \toprule[0.8pt]
        Entities & Vars. & Train & Test & Windows & Anom. \\
        \midrule[0.4pt]
        MSL P-15 & 55 & 3,682  & 2,856  & 2,797  & 80    \\
        SMAP T-3 & 25 & 2,876  & 8,579  & 8,520  & 302   \\
        SMD 1-1  & 38 & 28,479 & 28,479 & 28,420 & 3,166 \\
        \bottomrule[0.8pt]
    \end{tabular*}
\end{table}

\textbf{Datasets.} We evaluate \textsc{Gfm4Mtsad} on three representative MTSAD benchmarks spanning spacecraft telemetry and industrial server monitoring: Mars Science Laboratory (MSL P-15) and Soil Moisture Active Passive (SMAP T-3) \cite{hundman2018detecting}, and Server Machine Dataset (SMD 1-1) \cite{su2019robust}. Table~\ref{tab:datasets} summarizes the statistics of these evaluation series. Following standard evaluation protocols, we retain the original training and testing partitions and segment sequences into sliding windows of length $L = 60$ with a stride of $1$. A test window is labeled anomalous if it contains at least one anomalous timestamp. Within the anomaly-free training partition, the first $80\%$ of observations are used for model fitting and reference construction, while the remaining $20\%$ serve as a held-out validation block; sliding windows are constructed independently within each block to prevent boundary leakage. Feature standardization uses statistics fitted exclusively on the training block.

\textbf{Baselines.} We compare \textsc{Gfm4Mtsad} against nine representative baseline methods covering three technical categories: \textbf{(1)} generic contrastive and out-of-distribution (OOD) detection methods: SimCLR \cite{chen2020simple}, CSI \cite{tack2020csi}, and SSD \cite{sehwag2021ssd}; \textbf{(2)} temporal attention and discrepancy-based MTSAD methods: AnomalyTransformer \cite{xu2021anomaly}, DCdetector \cite{yang2023dcdetector}, and CTAD \cite{kim2023contrastive}; and \textbf{(3)} structure-aware and advanced contrastive MTSAD methods: CARLA \cite{darban2025carla}, CAROTS \cite{kim2025causality}, and PaAno \cite{park2026paano}. To demonstrate the generality of our adaptation framework across distinct foundation architectures, we instantiate \textsc{Gfm4Mtsad} on two pretrained GFMs—AnomalyGFM~\cite{qiao2025anomalygfm} and OwlEye~\cite{zheng2026owleye}—denoted as \textsc{Gfm4Mtsad}-A and \textsc{Gfm4Mtsad}-O, respectively. All methods share identical data partitions, window labels, and evaluation protocols, where timestamp-level baseline outputs are aggregated into window-level scores via the within-window maximum.

\textbf{Metrics.} Following recent mainstream studies~\cite{kim2025causality, darban2025carla, park2026paano}, three standard metrics without point adjustment are selected for evaluation: AUROC, AUPRC, and F1 Score. In practical anomaly-sparse scenarios such as \text{MSL P-15} spacecraft telemetry monitoring, low AUROC and AUPRC values indicate that true subsystem faults are buried beneath massive false alarms, rendering automated alerting impractical. Higher values across these metrics signify stronger fault discrimination and lower false-alarm overhead.

\textbf{Implementation Details.}
\textit{Training Protocol:}
We use a two-stage warm-start protocol. The temporal frontend is first trained for $30$ epochs on normal windows and synthetic views. Its weights initialize the joint model, whose task-specific adapters follow the initialization rules in Section~\ref{subsec:SICA}; the GFM and its native adaptation parameters load their
pretrained weights. We then optimize $\Theta$ jointly for $30$ epochs with a batch size of $256$, keeping the GFM backbone frozen.The MAPS reference mechanism remains fixed during joint training. Normal scoring references are refreshed after each epoch; when the graph prior is enabled, it is refitted using the updated frontend before recalculating scoring statistics. Reference fitting uses normal training windows in evaluation mode without gradient tracking. Checkpoints are evaluated every $5$ epochs and selected by the lowest joint validation loss on held-out normal windows and synthetic views. After restoring the selected checkpoint, inference references are refitted on normal training data and frozen. The main comparison uses three runs per entity--backbone configuration. The framework is implemented in PyTorch and evaluated on NVIDIA GeForce RTX 4090 D GPUs.

\textit{Parameters:} 
The DLinear frontend extracts $128$-dimensional temporal representations ($d_1$) per sensor using a moving-average kernel of size $25$, and the Token Bridge projects them to the native token dimension $d_z$ required by each GFM. IPRA employs a low-rank bottleneck with rank $r$ set to $16$, residual scaling factor $\gamma$ fixed at $1.0$, and zero-initialized up-projection weights $\mathbf{V}$. Graph construction sets the neighborhood size $k_g$ to $5$ per node. The prior weight $\lambda_A$ is set to $0.25$ for all AnomalyGFM and OwlEye settings. Optional directional screening retains a candidate only if, under a fixed reference scorer, the target sensor's response increases and its increase exceeds that of every other sensor. The main comparison uses $b_t$ for AnomalyGFM and the re-ranked score $s_t$ for OwlEye.
Optimization uses the Adam optimizer with a weight decay of $10^{-3}$, a base learning rate of $10^{-3}$ for temporal and adaptation modules, and $10^{-4}$ for lightweight GFM parameters $\psi$, scheduled via a $5$-epoch linear warm-up followed by cosine annealing and gradient-norm clipping at $0.5$. The contrastive temperature $T_c$ and pooling temperature $T_p$ are set to $0.2$ and $0.5$, respectively. The loss weights $\lambda_{\mathrm{n}}$, $\lambda_{\mathrm{r}}$, $\lambda_{\mathrm{s}}$, and $\lambda_{\mathrm{c}}$ are set to $1$, $0.2$, $1$, and $0.1$, respectively. TGEC fixes the default tail probability $\alpha_0$ at $0.05$, the fusion weight $\omega$ at $0.5$, and the calibrated evidence aggregation size $k_e$ at $2$. Our code is available at \href{https://github.com/yanzhuang-h/GFM4MTSAD}{https://github.com/yanzhuang-h/GFM4MTSAD}.

\begin{table*}[t]
\centering
\caption{Performance Comparison of MTSAD on Three Benchmark Datasets. The Best and Second-Best Results are Highlighted in \textbf{Bold} and \underline{Underlined} Text. AT Denotes AnomalyTransformer, G4M-A and G4M-O Denote \textsc{Gfm4Mtsad} Based on AnomalyGFM and OwlEye, Respectively.}
\label{tab:main-results}
\setlength{\tabcolsep}{3.6pt}
\renewcommand{\arraystretch}{1.12}
\begin{tabular}{c|c|ccc|ccc|ccc|cc}
\toprule
\multirow{2}{*}{Datasets} & \multirow{2}{*}{Metrics} & SimCLR & CSI & SSD & AT & DCdetector & CTAD & CARLA & CAROTS & PaAno & \textbf{\textsc{G4M}-A} & \textbf{\textsc{G4M}-O} \\
 & & (ICML20) & (NIPS20) & (ICLR21) & (ICLR22) & (KDD23) & (TKDE24) & (PR25) & (ICML25) & (ICLR26) & \textbf{(Ours)} & \textbf{(Ours)} \\
\midrule
\multirow{3}{*}{MSL P-15} & AUROC & 0.5246 & 0.4659 & 0.4621 & 0.3072 & 0.3909 & 0.5420 & 0.6956 & 0.6592 & 0.5104 & \underline{0.8747} & \textbf{0.9246} \\
 & AUPRC & 0.0280 & 0.0252 & 0.0255 & 0.0186 & 0.0215 & 0.0293 & \underline{0.4208} & 0.0378 & 0.0271 & 0.1731 & \textbf{0.6916} \\
 & F1 & 0.0947 & 0.0805 & 0.0712 & 0.0569 & 0.0562 & 0.1029 & \underline{0.5405} & 0.0950 & 0.0856 & 0.3318 & \textbf{0.8138} \\
\midrule
\multirow{3}{*}{SMAP T-3} & AUROC & 0.4956 & 0.5997 & 0.4543 & 0.5230 & 0.4717 & 0.5439 & 0.6654 & 0.4931 & \textbf{0.7923} & \underline{0.7233} & 0.7209 \\
 & AUPRC & 0.0650 & 0.0456 & 0.0691 & 0.2267 & 0.0382 & 0.0377 & 0.2841 & 0.0440 & 0.3908 & \underline{0.4120} & \textbf{0.4169} \\
 & F1 & 0.0814 & 0.0861 & 0.1189 & 0.3315 & 0.0799 & 0.0796 & 0.3897 & 0.0752 & \textbf{0.5505} & \underline{0.4922} & 0.4739 \\
\midrule
\multirow{3}{*}{SMD 1-1} & AUROC & 0.5400 & 0.4114 & 0.6010 & 0.5423 & 0.5078 & 0.5451 & 0.3901 & \underline{0.8740} & 0.5901 & \textbf{0.9089} & 0.8087 \\
 & AUPRC & 0.1127 & 0.0932 & 0.2559 & 0.2632 & 0.1155 & 0.1200 & 0.2430 & 0.5225 & 0.1213 & \underline{0.6050} & \textbf{0.6167} \\
 & F1 & 0.2211 & 0.2006 & 0.3095 & 0.3530 & 0.2008 & 0.2126 & 0.2676 & 0.5625 & 0.2976 & \underline{0.5680} & \textbf{0.6198} \\
\bottomrule
\end{tabular}
\end{table*}

\subsection{Performance Comparison (RQ1)}
\label{subsec:perf-comparison}

We present a comprehensive comparison between \textsc{Gfm4Mtsad} and representative baselines in Table~\ref{tab:main-results}, where \textsc{Gfm4Mtsad} secures the best or second-best performance across nearly all evaluation criteria. Both \textsc{Gfm4Mtsad}-A and \textsc{Gfm4Mtsad}-O substantially outperform competing methods, confirming that aligning temporal feature and relational contracts (\textbf{C1}) effectively unleashes pretrained GFMs for MTSAD. Specifically, \textsc{Gfm4Mtsad}-O excels on high-dimensional MSL via OwlEye's pattern dictionary and truncated attention reconstruction, which amplify acute relational deviations, whereas \textsc{Gfm4Mtsad}-A achieves stronger stability on SMAP and SMD through AnomalyGFM's dual normal--abnormal prototype alignment. Moreover, the two backbones differ markedly in native input dimensionality, as AnomalyGFM projects temporal representations into a substantially more compact node token space than OwlEye. This stronger token compression governs the granularity of temporal variations available to internal graph computations, while IPRA supplies complementary temporal information at the output stage. Coupled with differences in the backbones' native scoring mechanisms, these factors shape the ranking of anomalous windows and yield relative advantages across different datasets and evaluation metrics.

Compared with baselines across contrasting physical scenarios—namely aerospace telemetry (MSL and SMAP), characterized by sparse faults and abrupt command-driven transitions, and industrial server monitoring (SMD), featuring dense multi-metric co-fluctuations—\textsc{Gfm4Mtsad} demonstrates superior cross-scenario adaptability. Existing methods often favor a single regime; for instance, patch-based PaAno captures local temporal patterns in lower-dimensional SMAP but degrades under high-dimensional channel interactions in MSL and SMD. Conversely, \textsc{Gfm4Mtsad} generalizes robustly across both sparse telemetry and complex server workloads by jointly modeling channel-wise dynamics and cross-sensor couplings.

Compared with non-graph baselines—including instance-level contrastive methods (SimCLR, CSI, and SSD) and temporal discrepancy models (AnomalyTransformer, DCdetector, and CTAD)—as well as raw graph-based methods (CAROTS), \textsc{Gfm4Mtsad} establishes a more reliable topological modeling paradigm. Non-graph methods neglect inter-sensor topology and consequently suffer from high false alarms under benign non-stationary fluctuations, whereas relying on raw window-level graphs leaves CAROTS vulnerable to transient correlational noise. \textsc{Gfm4Mtsad} overcomes both limitations via SICA, anchoring message passing on persistent normal connectivity priors while leveraging IPRA to recover projection-excluded temporal details from $\mathbf{F}_t^\perp$.

Compared with advanced contrastive baselines (CARLA, CAROTS, and PaAno) that rely on heuristic augmentations or unconstrained negative sampling and often generate trivial outliers far from the true decision boundary, \textsc{Gfm4Mtsad} couples MAPS with TGEC to form a closed-loop supervision and calibration pipeline. Specifically, MAPS synthesizes residual-preserving interventions near the normal Graph-VAR boundary to provide informative training supervision, while TGEC selectively integrates multi-granularity evidence in the upper tail to sharpen fault rankings without distorting normal scores.

\begin{table*}[t]
\centering
\caption{SICA and MAPS Ablations with the OwlEye Backbone
(Native Window AUPRC and F1). Best and Second-Best Results
Are \textbf{Bold} and \underline{Underlined}, Respectively.}
\label{tab:ablation-joint}

\small
\setlength{\tabcolsep}{4pt}
\renewcommand{\arraystretch}{1.18}

\begin{tabularx}{\textwidth}{
    @{} l @{\hspace{10pt}}
    >{\raggedright\arraybackslash}X
    *{6}{>{\centering\arraybackslash}p{0.074\textwidth}} @{}
}
\toprule
\multicolumn{2}{l}{\multirow{2}{*}{Ablation Category and Variant}}
& \multicolumn{2}{c}{MSL-O}
& \multicolumn{2}{c}{SMAP-O}
& \multicolumn{2}{c}{SMD-O} \\
\cmidrule(lr){3-4}
\cmidrule(lr){5-6}
\cmidrule(lr){7-8}
& & AUPRC & F1 & AUPRC & F1 & AUPRC & F1 \\
\midrule

\multicolumn{2}{l}{\textbf{Full Model (SICA + MAPS)}}
& \textbf{0.8131} & \textbf{0.9182}
& \textbf{0.4519} & \underline{0.5283}
& \textbf{0.6932} & \textbf{0.6849} \\

\midrule
\multirow{4}{*}{SICA}
& w/ Minimal SICA + MAPS
& 0.6998 & 0.8439
& 0.3408 & 0.4215
& 0.2483 & 0.3232 \\

& w/o PCA Bridge Initialization
& 0.4203 & 0.6854
& 0.3928 & 0.4784
& \underline{0.6642} & \underline{0.6344} \\

& w/o Normal Graph Prior ($\lambda_A=0$)
& 0.6131 & 0.7526
& --- & ---
& --- & --- \\

& w/o IPRA Residual Bypass
& 0.5723 & 0.8066
& 0.3410 & 0.4621
& 0.4756 & 0.5506 \\

\midrule
MAPS & w/ Random Direct-Edit Selection
& \underline{0.7555} & \underline{0.8742}
& \underline{0.4277} & \textbf{0.5424}
& 0.6445 & 0.6175 \\

\bottomrule
\end{tabularx}

\vspace{3pt}
\parbox{\textwidth}{\footnotesize
\textit{Note:} ``---'' denotes an inactive normal graph prior.
All variants share the same warm-start frontend within each entity.}

\end{table*}

\begin{table}[t]
\centering
\caption{Ablation Study of TGEC Scoring Variants Across Six Configurations (AUPRC), where Act.\ Denotes the Test Tail Activation Rate.}
\label{tab:ablation-TGEC}
\setlength{\tabcolsep}{3.8pt}
\renewcommand{\arraystretch}{1.12}
\begin{tabular}{l|ccccc|r}
\toprule
Config. & Native & ECDF & Global & Direct-only & \textbf{TGEC} & Act. \\
\midrule
SMD-A & \textbf{0.6275} & \underline{0.6274} & 0.5210 & 0.5198 & 0.5217 & 24.07\% \\
SMAP-A & \textbf{0.4942} & 0.3835 & 0.4096 & 0.3483 & \underline{0.4112} & 22.58\% \\
MSL-A & 0.4775 & 0.4795 & \textbf{0.7070} & \underline{0.5709} & \underline{0.5709} & 1.11\% \\
SMD-O & 0.6932 & 0.6894 & \underline{0.6970} & 0.6947 & \textbf{0.7015} & 14.45\% \\
SMAP-O & \textbf{0.4519} & 0.4211 & 0.4424 & 0.3440 & \underline{0.4504} & 9.28\% \\
MSL-O & \underline{0.8131} & \textbf{0.8366} & 0.1230 & \textbf{0.8366} & \textbf{0.8366} & 0.00\% \\
\bottomrule
\end{tabular}
\end{table}

\subsection{Ablation Study (RQ2)}
\label{subsec:ablation}

We evaluate the contributions of SICA, MAPS, and TGEC in Tables~\ref{tab:ablation-joint} and~\ref{tab:ablation-TGEC}. Training-stage ablations share the same warm-start frontends, matched downstream initializations, data ordering, and supervision budgets over 30 epochs. Table~\ref{tab:ablation-joint} reports native window AUPRC and F1 to isolate representation learning effects, whereas Table~\ref{tab:ablation-TGEC} compares TGEC scoring variants on identical frozen Full checkpoints.

\textit{Effectiveness of SICA.}
Our method benefits from both the complete alignment architecture and its individual contractual components. Replacing SICA with a minimal dimension-compatible interface consistently reduces both AUPRC and F1 across all evaluated entities, causing a particularly pronounced degradation on SMD. Similarly, removing either PCA-based Token Bridge initialization or the IPRA residual bypass degrades both metrics across all entities, confirming the necessity of anchoring initial token projections within the normal feature covariance subspace and recovering projection-excluded temporal details. Furthermore, incorporating the normal graph prior substantially strengthens detection performance on high-dimensional MSL in both AUPRC and F1 by suppressing transient correlational noise. Together, these results demonstrate the complementary contributions of feature alignment, residual information preservation, and normal connectivity priors.

\textit{Effectiveness of MAPS.}
We isolate the contribution of candidate selection using \textit{w/ Random Direct-Edit Selection}. This variant retains mechanism-based candidate generation, validity constraints, context propagation, and supervision budgets, while replacing boundary-proximity ranking with uniform sampling among valid direct-edit candidates. Across MSL, SMAP and SMD, boundary-ranked selection consistently outperforms uniform selection in AUPRC and yields further F1 improvements on MSL and SMD. These results confirm that prioritizing hard, informative interventions near the normal mechanism boundary provides stronger discriminative supervision than unranked sampling within the same candidate pool.

\begin{figure}[t]
    \centering
    \includegraphics[width=\columnwidth]{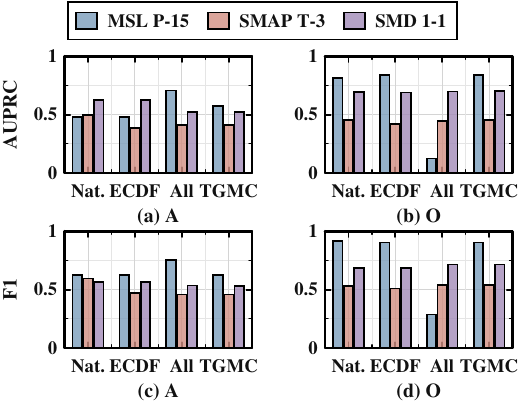}
    \caption{TGEC scoring ablation: top, AUPRC; bottom, raw oracle F1. A/O denote \textsc{Gfm4Mtsad}-A/O. Nat.: native scores; ECDF: normal-reference scoring; All: global consensus; TGEC: tail consensus.}
    \label{fig:TGEC-scoring-ablation}
\end{figure}

\textit{Effectiveness of TGEC.}
Table~\ref{tab:ablation-TGEC} compares five scoring variants on identical frozen Full checkpoints. TGEC restricts evidence fusion to the train-defined upper tail, maintaining exact central identity where $s_t$ equals $u_t$ below $\tau_0$ and preserving the ECDF ordering within the normal score body. On SMD and SMAP, TGEC consistently outperforms both \textit{Global} consensus and \textit{Direct-only} tail scoring, confirming the synergistic value of tail restriction and multi-granularity evidence fusion. The necessity of protecting the normal body against unconstrained modification is particularly evident on MSL-O: when no test windows exceed the tail gate, TGEC faithfully retains the strong ECDF ranking performance, whereas global fusion causes severe performance degradation by distorting benign rankings. Together, these results validate selective tail re-ranking for sharpening anomaly detection while safeguarding the central score structure.

\subsection{Mechanism Analysis (RQ3)}
\label{subsec:mechanism-analysis}

\begin{figure}[t]
    \centering
    \includegraphics[width=\columnwidth]{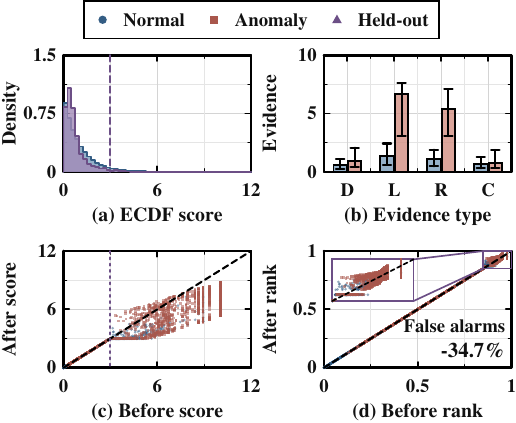}
    \caption{TGEC on SMD (\textsc{Gfm4Mtsad}-O): (a) normal distributions and tail gate; (b) evidence medians and IQRs (D/L/R/C: direct/local/relation/context); (c) score mapping; (d) rank changes with the upper tail enlarged (both axes: 0.85--1). Normal/anomalous tail windows: 2,010/2,097. At the normal-q99 threshold, false positives decrease from 918 to 599 (34.7\%).}
    \label{fig:TGEC-mechanism}
\end{figure}

To provide an intuitive understanding of how \textsc{Gfm4Mtsad} refines anomaly decisions at inference time, we visualize the internal calibration mechanics of TGEC on SMD 1-1 (\textsc{Gfm4Mtsad}-O) in Fig.~\ref{fig:TGEC-mechanism}. Specifically, we examine how the framework separates genuine faults from benign operational spikes across four sequential stages: tail gating, multi-granularity evidence decoupling, score transformation, and rank calibration.

As shown in Fig.~\ref{fig:TGEC-mechanism}(a) and (c), the empirical normal and held-out validation distributions concentrate densely below the train-defined tail threshold $\tau_0$, where TGEC enforces an exact identity mapping along the diagonal so that routine normal fluctuations remain completely undisturbed. Once scores cross $\tau_0$ into the active upper tail, relying solely on baseline surprisal becomes prone to false alarms caused by transient single-sensor spikes. However, examining the calibrated multi-granularity evidence in Fig.~\ref{fig:TGEC-mechanism}(b) reveals a clear structural separation: while benign normal tail windows and true anomalies exhibit comparable direct (D) and contextual (C) responses, true anomalies induce markedly stronger local (L) and relational (R) evidence due to underlying inter-sensor dependency breakdowns. By selectively integrating these complementary cues above $\tau_0$, the score mapping in Fig.~\ref{fig:TGEC-mechanism}(c) and the zoomed-in tail trajectory in Fig.~\ref{fig:TGEC-mechanism}(d) effectively elevate genuine anomalous windows while attenuating isolated normal spikes. Consequently, under the fixed normal-training $99\text{th}$-percentile threshold rule, TGEC substantially reduces false-alarm windows while preserving true anomaly recall, confirming its interpretability and practical value for reliable alarm prioritization.

\subsection{Parameter Study (RQ4)}
\label{subsec:param-study}

\begin{figure}[t]
    \centering
    \includegraphics[width=\columnwidth]{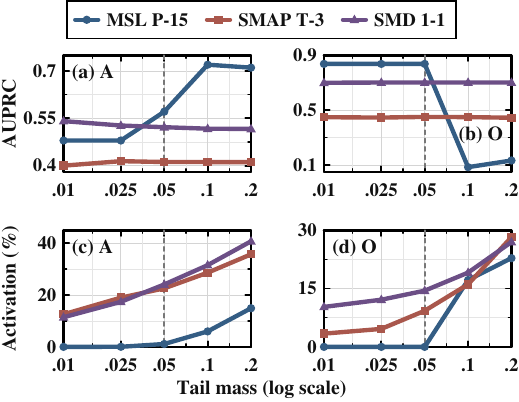}
    \caption{Tail-mass sensitivity: top, AUPRC; bottom, test-window tail activation. A/O denote \textsc{Gfm4Mtsad}-A/O. The fusion weight is fixed at 0.5. Dashed lines mark the default tail mass of 0.05. Horizontal axes are logarithmic; vertical ranges vary by panel.}
    \label{fig:TGEC-tail-mass}
\end{figure}

\begin{figure}[t]
    \centering
    \includegraphics[width=\columnwidth]{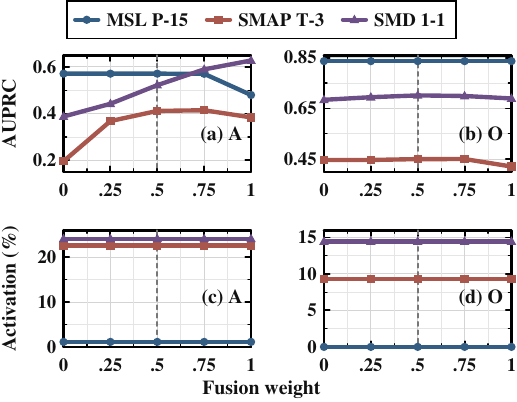}
    \caption{Fusion-weight sensitivity: top, AUPRC; bottom, test-window tail activation. A/O denote \textsc{Gfm4Mtsad}-A/O. The tail mass is fixed at 0.05; dashed lines mark the default fusion weight of 0.5. Vertical ranges vary by panel.}
    \label{fig:TGEC-fusion-weight}
\end{figure}

\begin{figure}[t]
    \centering
    \includegraphics[width=\columnwidth]{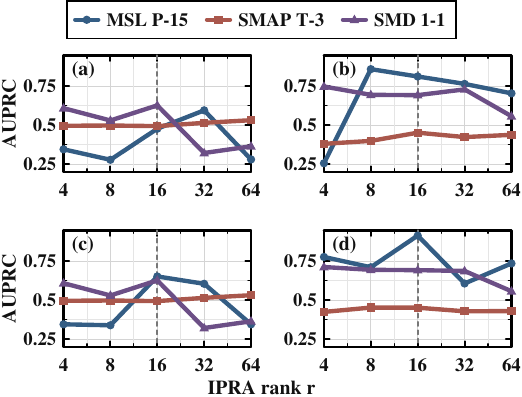}
    \caption{IPRA rank sensitivity: native AUPRC at validation-selected (top) and final (bottom, epoch 30) checkpoints. Left/right: \textsc{Gfm4Mtsad}-A/O. Dashed lines mark the default rank of 16. All panels share the vertical scale; ranks are spaced logarithmically.}
    \label{fig:ipra-rank}
\end{figure}

We examine three hyperparameters of \textsc{Gfm4Mtsad} across the same six entity--backend configurations. For TGEC, we vary $\alpha_0\in\{0.01,0.025,0.05,0.10,0.20\}$ while $\omega$ is fixed at $0.5$, and vary $\omega\in\{0,0.25,0.5,0.75,1\}$ while $\alpha_0$ is fixed at $0.05$. The approximately geometric tail-mass grid spans the upper 1\%--20\% of the normal reference, covering a twentyfold change in selective re-ranking scope; the uniform fusion grid covers the full mixture interval, including both endpoints. These sweeps reuse the validation-selected checkpoints and frozen normal references. For IPRA, $r\in\{4,8,16,32,64\}$ doubles the compensation bottleneck at each step under the same warm-start and 30-epoch joint-training protocol.

\textit{Impact of Tail Probability Mass $\alpha_0$.} Increasing $\alpha_0$ lowers the gate and monotonically expands test-window activation (Fig.~\ref{fig:TGEC-tail-mass}). This relationship need not be linear, because activation depends on the test-score distribution. Across the full grid, AUPRC varies by at most 0.0249 on SMD and SMAP. MSL illustrates the role of selective gating: broader tails benefit MSL-A, whereas MSL-O retains its highest AUPRC over tail masses from 0.01 to 0.05. Thus, the gate controls where evidence fusion acts, separating tail re-ranking from preservation of the score body.

\textit{Impact of Evidence Fusion Weight $\omega$.} Varying $\omega$ leaves activation unchanged while adjusting rankings within the selected tail (Fig.~\ref{fig:TGEC-fusion-weight}). A weight of one recovers ECDF-only scoring; a weight of zero uses the auxiliary consensus subject to the tail floor. Intermediate weights improve AUPRC over both endpoints on SMAP with either backend and on SMD-O. SMD-A instead improves as the baseline contribution increases, while MSL-O remains constant because its gate is inactive. These responses distinguish the gate's selection role from the fusion weight's ranking role. Activation denotes gate membership, including when the fusion weight is one and scores are unchanged.

\textit{Impact of IPRA Bottleneck Rank $r$.} The rank controls the capacity of IPRA’s residual compensation pathway. Fig. (\ref{fig:ipra-rank}) compares validation-selected checkpoints with the final checkpoints after 30 epochs. Across the six entity–backend configurations, a rank of 16 achieves the highest mean native AUPRC under both conditions, reaching 0.5929 and 0.6392, respectively. Increasing the rank to 32 or 64 enlarges the IPRA parameter budget linearly without improving these averages. These results support a moderate bottleneck capacity and motivate retaining 16 as the shared default rank, balancing detection performance with compact residual compensation.

\section{Conclusion and Future Work}
\label{sec:conclusion_and_future_work}

This paper presented \textsc{Gfm4Mtsad}, an adaptation framework that unleashes pretrained GFMs for unsupervised MTSAD. Within the framework, SICA harmonizes temporal representations and sensor topologies through feature, structure, and information contracts while recovering projection-excluded details via IPRA. MAPS furnishes targeted supervision through residual-preserving mechanism interventions near the normal boundary, and TGEC calibrates multi-granular evidence against normal operational references to refine tail anomaly candidates while preserving benign score rankings. Extensive experiments with AnomalyGFM and OwlEye backbones across MSL P-15, SMAP T-3, and SMD 1-1 demonstrate the effectiveness of our approach, where \textsc{Gfm4Mtsad} achieves superior detection precision and ablations confirm the necessity of each contractual and calibration design. These findings establish that coordinated temporal and relational adaptation enables pretrained graph models to serve as effective anomaly detection backbones.

Future work will investigate how native token capacity and pretraining inductive biases govern the transferability of heterogeneous GFMs, motivating capacity-aware adaptation strategies that dynamically coordinate representation granularity across diverse foundation architectures. Beyond adapting off-the-shelf graph backbones, we plan to explore unified foundation pretraining across time series and graphs to co-learn continuous temporal dynamics and discrete topological priors within a shared latent space. A further direction is continual adaptation under non-stationary operating regimes and distribution shifts, enabling structural priors and tail calibration references to evolve online while preserving pretrained foundation knowledge.

\bibliographystyle{IEEEtran}
\bibliography{references}

\newpage

\end{document}